\documentclass[10pt,twocolumn,letterpaper]{article}

 \usepackage{cvpr}              

\usepackage[dvipsnames]{xcolor}
\usepackage[normalem]{ulem}

\definecolor{cvprblue}{rgb}{0.21,0.49,0.74}
\usepackage[pagebackref,breaklinks,colorlinks,citecolor=cvprblue]{hyperref}

\usepackage{tikz}
\usetikzlibrary{intersections, calc, decorations.pathreplacing, decorations.markings}
\usepackage{listings}

\lstdefinestyle{mystyle}{
    basicstyle=\ttfamily\footnotesize,
    backgroundcolor=\color{lightgray!20},
    commentstyle=\color{gray},
    keywordstyle=\color{blue},
    stringstyle=\color{orange},
    breaklines=true,
    frame=single,
    captionpos=b,
    keepspaces=true,
    numbers=left,
    numberstyle=\tiny\color{gray},
    numbersep=5pt,
    showspaces=false,
    showstringspaces=false,
    showtabs=false,
    tabsize=2
}
\def\paperID{*****} 
\def\confName{CVPR}
\def\confYear{2024}

\definecolor{darkgreen}{RGB}{30,150,30}
\definecolor{darkblue}{RGB}{0,0,127}
\definecolor{darkyellow}{RGB}{171,133,0}
\definecolor{darkred}{RGB}{180,20,20}
\definecolor{darkmagenta}{RGB}{127,0,127}
\definecolor{darkcyan}{RGB}{0,127,127}
\definecolor{chromeyellow}{rgb}{1.0, 0.65, 0.0}
\definecolor{amber}{rgb}{1.0, 0.75, 0.0}

\newif\ifdrafting
 \draftingfalse 

\ifdrafting
  \newcommand{\OG} [1] {\textcolor{darkgreen}{[OG: #1]}}
  \newcommand{\HS} [1] {\textcolor{darkred}{[HS: #1]}}
  \newcommand{\AB} [1] {\textcolor{darkmagenta}{[AB: #1]}}
  \newcommand{\DL} [1] {\textcolor{darkyellow}{[DL: #1]}}
  \newcommand{\TODO} [1] {{\color{darkcyan}{\bf [TODO: #1]}}}

\else
  \newcommand{\OG} [1] {}
  \newcommand{\HS} [1] {}
  \newcommand{\AB} [1] {}
  \newcommand{\DL} [1] {}
  \newcommand{\TODO} [1] {}
  
\fi

\title{nvTorchCam: {A}n Open-source Library for Camera-Agnostic Differentiable Geometric Vision}

\author{Daniel Lichy$^{\dagger,\diamond}$ \hspace{3mm}
Hang Su$^{\diamond}$ \hspace{3mm} 
Abhishek Badki$^{\diamond}$ \hspace{3mm}
Jan Kautz$^{\diamond}$ \hspace{3mm}
Orazio Gallo$^{\diamond}$\\
$^{\dagger}$University of Maryland \hspace{3mm} $^{\diamond}$NVIDIA\\
}

\begin{document}
\maketitle
\begin{abstract}
We introduce \texttt{nvTorchCam}, an open-source library under the Apache 2.0 license, designed to make deep learning algorithms camera model-independent. \texttt{nvTorchCam} abstracts critical camera operations such as projection and unprojection, allowing developers to implement algorithms once and apply them across diverse camera models—including pinhole, fisheye, and 360 equirectangular panoramas, which are commonly used in automotive and real estate capture applications. Built on PyTorch, \texttt{nvTorchCam} is fully differentiable and supports GPU acceleration and batching for efficient computation. Furthermore, deep learning models trained for one camera type can be directly transferred to other camera types without requiring additional modification. In this paper, we provide an overview of \texttt{nvTorchCam}, its functionality, and present various code examples and diagrams to demonstrate its usage. Source code and installation instructions can be found on the \href{https://github.com/NVlabs/nvTorchCam}{\texttt{nvTorchCam}} GitHub page.
\end{abstract} 
\section{Introduction}

\begin{figure}[h]
    \centering
    \includegraphics[width=\linewidth]{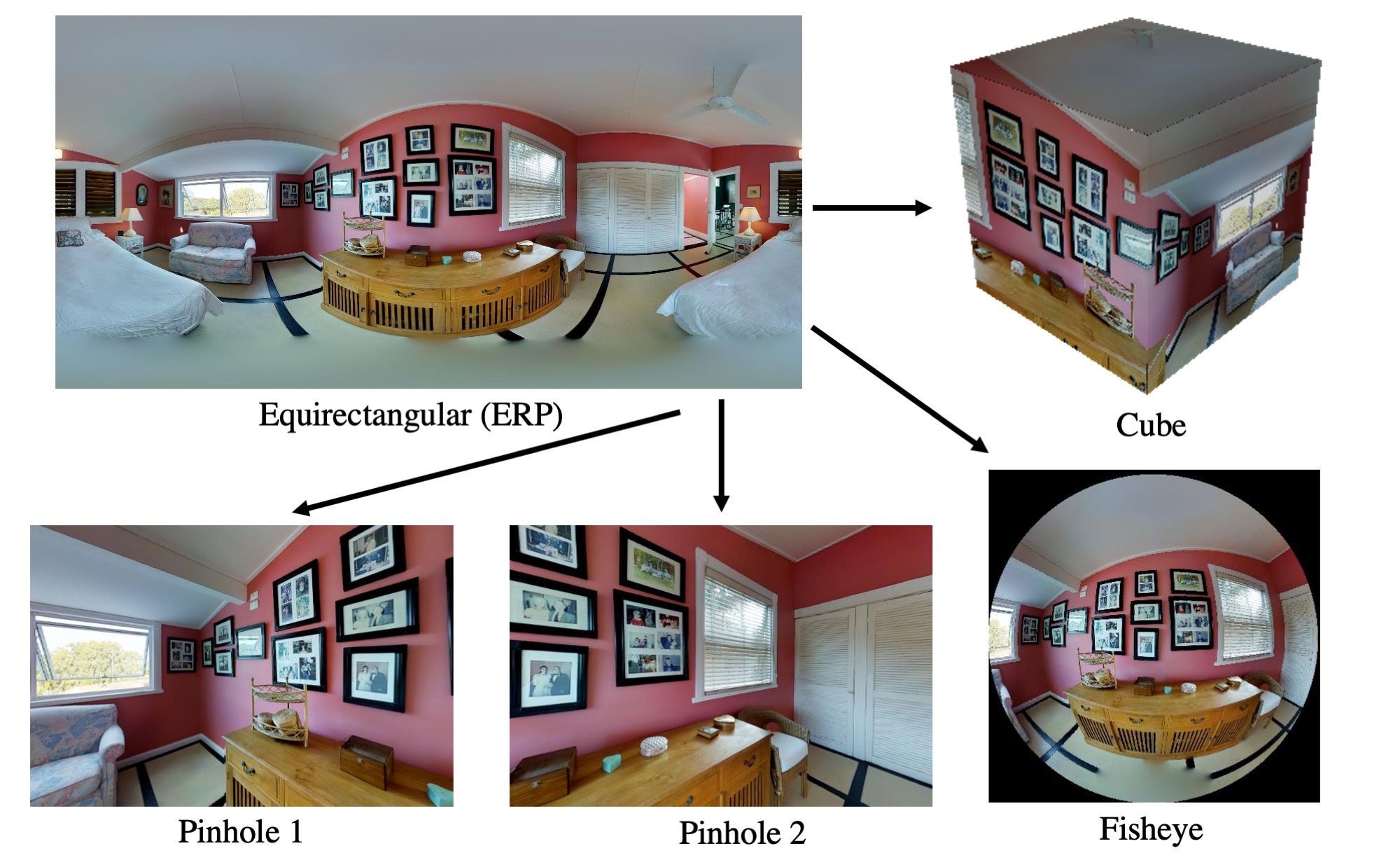}
    \caption{Shows an equirectangular panorama (ERP) being resampled to a cubemap, two pinhole images with different extrinsic rotations, and a fisheye projection.}
    \label{fig:resample_by_intrinsics}
\end{figure}

Computer vision applications often rely on diverse camera models, such as pinhole, fisheye, and 360 equirectangular panoramas (ERP), each with unique characteristics that influence image processing and algorithm performance. For example, cameras with larger fields of view (FoV) can capture more of the scene at once, which is particularly useful in automotive and real estate applications, but introduce distortions that must be accounted for during processing. Handling different camera models traditionally requires implementing model-specific solutions, which increases complexity and limits flexibility. To address these challenges, we introduce \textit{nvTorchCam}, an open-source library designed to abstract camera models and enable differential camera operations. With \textit{nvTorchCam}, developers can implement algorithms once and apply them to a variety of camera models seamlessly, leveraging GPU acceleration and differentiability through PyTorch.

\textit{nvTorchCam} achieves this by defining an abstract base class, \texttt{CameraBase}, that standardizes core operations like point-to-pixel projection and pixel-to-ray conversion. For example, a deep-network developer would only need to ensure that their dataloader returns a subclass of \texttt{CameraBase}. If they are working with a dataset of pinhole images, the dataloader would return a \texttt{PinholeCamera} object; if the developer later switches to a fisheye dataset, they would simply return a \texttt{FisheyeCamera} object, another subclass of \texttt{CameraBase}. Since both camera types inherit from this base class, the network code remains unchanged---handling data from different camera models seamlessly. This object-oriented design abstracts away the specifics of each camera model, allowing algorithms to work across different camera models without modification. Additionally, \textit{nvTorchCam} simplifies the bug-prone task of managing camera parameters during data augmentations, such as cropping and flipping, ensuring accurate transformations regardless of the camera model.


A critical class of algorithms for which \textit{nvTorchCam} provides camera-agnostic implementations is backward warping. These algorithms enable a wide range of tasks, such as warping between different camera models, rectification, plane and sphere sweeping, and depth consistency checking. The \textit{Kornia} library~\cite{kornia}, one of the inspirations for this work, offers limited backward warping capabilities through its \texttt{warp\_frame\_depth} function but only supports standard pinhole camera models. In contrast, \textit{nvTorchCam} generalizes this operation to support any camera model that extends the \texttt{CameraBase} class, for example, large-FoV formats such as fisheye and ERP cameras. Furthermore, \textit{nvTorchCam} introduces a unique mechanism for tracking invalid points in an image---such as points on the edge of a fisheye image that don't correspond to valid rays or 3D points located behind the camera---which is crucial for achieving accurate backward warping in complex camera setups. We applied this backward warping mechanism extensively in our project, FoV-Depth~\cite{Fova-depth}, which proposes a solution for multiview stereo depth estimation that can be trained on widely available pinhole datasets and generalized to large-FoV datasets.

The remainder of this paper is structured as follows: First, we review several popular libraries that address similar challenges, highlighting the gaps that \textit{nvTorchCam} is designed to fill. Next, we provide an overview of the library's structure. This is followed by an in-depth exploration of the mathematical foundations that the library models, accompanied by detailed code examples and diagrams to guide its use. Finally, we offer concluding remarks and discuss potential future developments. We also include a reference appendix with detailed mathematical explanations of the camera projection models currently implemented in the library.    
\section{Prior Work}

\textbf{Traditional CPU-Based Libraries}: Existing CPU libraries, such as COLMAP~\cite{Colmap} and OpenCV~\cite{opencv}, have served as an inspiration for \textit{nvTorchCam}'s camera model implementations. COLMAP, in particular, offers classes for different cameras that laid the groundwork for \textit{nvTorchCam}'s design. However, these libraries lack GPU support, cannot be easily integrated into deep learning frameworks, and offer no support for cameras with fields of view (FoV) greater than 180 degrees or non-central cameras. Furthermore, they do not provide differentiable operations, which are essential for modern deep learning pipelines.

\textbf{Deep Learning Libraries}: Kornia~\cite{kornia} offers some functionality that is similar to \textit{nvTorchCam}. However, it does not provide unified, interchangeable camera models for generic implementations. For example, Kornia’s functional interface for camera distortion makes it difficult to implement generic operations across different camera models. A notable example is Kornia’s \texttt{depth\_to\_3d} function, which only supports unprojecting a depth map to a 3D point cloud for pinhole cameras. In contrast, \textit{nvTorchCam} cameras have a \texttt{unproject\_depth} function that does the same operation for any camera model. Similarly, Kornia’s \texttt{warp\_frame\_depth} function is limited to pinhole cameras, whereas \textit{nvTorchCam}’s \texttt{backward\_warp} function supports warping operations between different camera models, allowing for advanced tasks such as resampling from one central camera model to another, undistorting a fisheye or ERP image to one or multiple pinhole images, and constructing cost-volumes.

PyTorch3D~\cite{pytorch3d} offers projection and unprojection functions and limited fisheye image support, but as a rasterization-focused library, its primary design is not centered around general camera operations. It lacks essential warping operations, support for ERP cameras, and functionality for adjusting camera intrinsics during operations like cropping and flipping. Additionally, its complexity makes it challenging to add new camera models.

NerfStudio~\cite{nerfstudio2023}, on the other hand, is optimized for Neural Radiance Fields (NeRFs), which operate primarily on rays. While it provides operations for unprojection (converting pixels to ray directions), it lacks the corresponding projection functions that are necessary for backward warping operations.
   
\textbf{Project-Specific Implementations}: Many deep learning projects that require camera models, such as MVSNet~\cite{MVSNet}, OmniMVS~\cite{won2020end}, and MODE~\cite{Li_Jin2022MODE}, rely on one-off implementations tailored to specific tasks or camera setups. This approach hinders code sharing and reuse, leading to incompatible development efforts. A striking example of this limitation, which \textit{nvTorchCam} aims to address, is the comparison between casMVSNet~\cite{CasMVSNet} and 360MVSNet~\cite{360MVSNet}. Although these two methods share nearly identical methodologies, they differ in their camera model implementations: casMVSNet is designed for pinhole cameras, while 360MVSNet focuses on equirectangular panorama (ERP) cameras. With \textit{nvTorchCam}, these two methods can seamlessly share the same implementation, streamlining development and fostering code reuse across different models.

\section{Library Structure}

The \textit{nvTorchCam} library is organized into several key components, each serving a specific role in the camera processing pipeline:

\begin{itemize}
    \item \textbf{cameras:} This module defines camera objects that encapsulate the parameters of different camera models. It provides essential methods for projection and conversion from pixel-to-rays.

    \item \textbf{cameras\_functional:} A functional interface offering camera projection and pixel-to-ray operations without maintaining camera state. Similar to how \texttt{torch.nn.functional} compares to \texttt{torch.nn}, all camera parameters are passed directly to the functions rather than being stored in an object, providing greater flexibility.
    \item \textbf{utils:} A collection of functions for common image operations. This includes interpolation, normalizing and unnormalizing image coordinates and intrinsic matrices, as well as applying transformations to points and vectors. 

    \item \textbf{warpings:} This module includes backward warping functions and their derivatives, such as resampling by intrinsics, stereo rectification, cropping, and affine transformations. It also supports plane and sphere sweeping techniques—commonly known as cost-volume construction—frequently used in multiview stereo depth estimation. A number of examples can be found in Section~\ref{sec:warping}.

    \item \textbf{diff\_newton\_inverse:} A utility class designed for iterative computation of inverse functions, commonly used in tasks such as undistortion, or pixel-to-ray conversion. For mathematical details, see Section~\ref{subsec:differentiable_newton_inverse}.
    \OG{Why does this have a link to the paragraph but the other items don't? If possible add to all.}
    \DL{@OG There is not a section for each one of these points. This differentiable Newton inverse is something on an easter egg/intellectual exercise for me. Actually someone on Lei's team asked me about differentiablity in the library. And I said this iterative undistortion wasn't not, but you can probably make it so. This just explains how to do it with a little math. Nevertheless, I added a reference to the warping point above.}
\end{itemize}
\section{Preliminary Notation}

\subsection{Doc-string Formatting and Inferred Batch Size}

Most functions in \textit{nvTorchCam} are designed to intelligently infer and handle tensor dimensions that resemble batches, allowing for efficient manipulation of data in both singular and batched forms. This capability and how it is documented is best illustrated through an example:
consider the doc-string for the function \texttt{utils.apply\_matrix}, which transforms batches of point groups using batches of matrices:

\begin{lstlisting}[language=Python, caption={Example doc-string for \texttt{apply\_matrix}}]
def apply_matrix(A: Tensor, pts: Tensor) -> Tensor:
    """ Transform batches of groups of points by batches of matrices
    Args:
        A: (*batch_shape, d, d)
        pts: (*batch_shape, *group_shape, d) 
    Returns:
        : (*batch_shape, *group_shape, d)
    """
\end{lstlisting}

The shape \texttt{(*batch\_shape, d, d)} indicates that the matrix \texttt{A} must have at least two dimensions representing rows and columns, while any additional preceding dimensions are interpreted as batch dimensions. Similarly, the shape \texttt{(*batch\_shape, *group\_shape, d)} for \texttt{pts} implies that the leading dimensions are batch-like and should match the batch dimensions of \texttt{A}, the final dimension of \texttt{pts} matches the last dimension of \texttt{A}, ensuring compatibility with the matrix dimensions, and between these there could be any number of dimensions where each point in a group is multiplied by the same matrix. Example shapes are shown in Table \ref{tab:infered_batch}. 
Most other functions and camera objects in the library follow a similar approach to inferred batching and are documented accordingly.

\begin{table*}[h]
\centering
\begin{tabular}{|c|c|p{8cm}|}
\hline
\textbf{Input Shapes} & \textbf{Output Shape} & \textbf{Description} \\ \hline
\texttt{(3,3)} and \texttt{(3,)} & \texttt{(3,)} & Multiply a single vector by a single matrix. \\ \hline
\texttt{(3,3)} and \texttt{(10,3)} & \texttt{(10,3)} & Multiply a group of 10 vectors by a single matrix. \\ \hline
\texttt{(12,6,4,3)} and \texttt{(12,6,4,3)} & \texttt{(12,6,4,3)} & Multiply a group of \texttt{(12,6,4)} vectors by a single matrix. \\ \hline
\texttt{(4,3,3)} and \texttt{(4,3)} & \texttt{(4,3)} & Multiply a batch of 4 vectors by a batch of 4 matrices. \\ \hline
\texttt{(5,4,3,3)} and \texttt{(5,4,10,3)} & \texttt{(5,4,10,3)} & Multiply a batch of \texttt{(5,4)} groups of 10 vectors by a batch of \texttt{(5,4)} matrices. \\ \hline
\end{tabular}
\caption{Examples of input and output shapes for \texttt{apply\_matrix} function.}
\label{tab:infered_batch}
\end{table*}

\subsection{Mathematical Notations}

If we have a function $f: A \rightarrow B \times C$, we use subscripts to denote the two components, i.e., $f_1: A \rightarrow B$ and $f_2: A \rightarrow C$.

The cube is defined as: 
\[
\mathbb{C} = \{x \in \mathbb{R}^3 : \|x\|_\infty = 1\}
\]

The sphere is defined as: 
\[
\mathbb{S}^2 = \{x \in \mathbb{R}^3 : \|x\|_2 = 1\}
\]

\section{Cameras}

In this section, we introduce our mathematical model of a camera. We then show how this is implemented in the library via camera objects, and how to create and manipulate these objects.

\subsection{Mathematical Model of Cameras}
\label{sec: camera_model_defs}
We define a camera in a manner similar to FoVA-Depth~\cite{Fova-depth} but extend the concept to support non-central cameras.

\textbf{General Cameras:} We define a general camera as a tuple $(U, A, \phi, p)$, where:

\begin{itemize}
    \item $U$ represents a 2D surface, referred to as the sensor. In \textit{nvTorchCam}, $U$ is either subset of $\mathbb{R}^2$ or the unit cube.
    
    \item $A$ denotes the valid regions of space that project onto $U$. This set also allows us to restrict regions behind the camera, typically specified by a minimum depth (`z\_min') or a minimum distance (`dist\_min').
    
    \item $\phi: U \rightarrow \mathbb{R}^3 \times S^2$ is the pixel-to-ray function, which maps pixels on the sensor to rays in 3D space.
    
    \item $p: A \rightarrow U \times \mathbb{R}$ is the projection function, which projects points in $A$ onto the camera sensor and returns their corresponding distances.
\end{itemize}

To ensure consistency between projection and the pixel-to-ray function, we require that for all $x \in A$:
\begin{equation}
    x = \phi_1(p_1(x)) + p_2(x)\phi_2(p_1(x))
    \label{project_unproject_consistency}
\end{equation}
Cameras with these properties are represented by classes extending the class \texttt{CameraBase}. We will get into the specifics of this when we discuss camera usage in the library.

\textbf{Affine Cameras:} We define affine cameras as those with $U \subseteq \mathbb{R}^2$, that have a final affine transform as the last step of projection. In other words their projection functions $p$ can be decomposed as:
\begin{equation}
\begin{bmatrix} 
p(x) \\ 
1 
\end{bmatrix} 
= 
\begin{bmatrix} 
f_1 & 0 & c_1 \\ 
0 & f_2 & c_2 \\ 
0 & 0 & 1 
\end{bmatrix} 
\begin{bmatrix} 
\hat{p}(x) \\ 
1 
\end{bmatrix},
\end{equation}
where, $f_1$ and $f_2$ are usually interpreted as focal lengths, and $(c_1,c_2)$ as the principal points. Cameras with this structure inherit from the \texttt{TensorDictionaryAffineCamera} class, which allows for easy adjustment of the camera parameters when cropping images.

\textbf{Central Cameras:} A central camera is a special case where all captured rays arrive at a single point, which is considered the origin.
In mathematical terms, a camera $(U, A, \phi, p)$ is central if $\phi_1(u) = 0$ for all $u \in U$. For these cameras \textit{nvTorchCam} provides additional operations, such as \texttt{resample\_by\_intrinsics}.
Cameras have a function \texttt{is\_central} which returns \texttt{True} if the camera is central and \texttt{False} otherwise.



\subsection{Camera Models and Their Usage}

In this section, we enumerate the camera models supported by \textit{nvTorchCam}, discuss how they implement the general definition of a camera, and explain how to use their fundamental functions.

\subsubsection{Supported Models}
The currently supported camera models include:
\begin{itemize}
    \item \texttt{PinholeCamera}
    \item \texttt{OrthographicCamera}
    \item \texttt{OpenCVCamera}
    \item \texttt{EquirectangularCamera}
    \item \texttt{OpenCVFisheyeCamera}
    \item \texttt{BackwardForwardPolynomialFisheyeCamera}
    \item \texttt{Kitti360FisheyeCamera}
    \item \texttt{CubeCamera}
\end{itemize}
The mathematical details of their projection models can be found in appendix~\ref{sec:projection_details} and on the GitHub page.

\subsubsection{Creating Cameras}

Cameras are instantiated via their static \texttt{make} method, as shown in the example below:
\begin{lstlisting}[language=Python, caption={Creating cameras in \textit{nvTorchCam}}]
>>> import nvtorchcam.cameras as cameras
>>> pin_cam = cameras.PinholeCamera.make(torch.eye(3))
>>> ortho_cam = cameras.OrthographicCamera.make(torch.eye(3))
\end{lstlisting}
Cameras can also be created with arbitrary batch shapes:
\begin{lstlisting}[language=Python, caption={Creating a batch of orthographic cameras}]
>>> intrinsics = torch.eye(3).reshape(1,1,3,3).expand(2,4,3,3)
>>> ortho_cam = cameras.OrthographicCamera.make(intrinsics)
>>> ortho_cam.shape
torch.Size([2, 4])
\end{lstlisting}

\subsubsection{Essential Camera Functions}

The two most essential functions for camera models in \textit{nvTorchCam} are \texttt{project\_to\_pixel} and \texttt{pixel\_to\_ray}, corresponding to $p$ and $\phi$, respectively, in the general camera definition.

\paragraph{Function: project\_to\_pixel}
The \texttt{project\_to\_pixel} function maps 3D points $(x, y, z)$ in space to their corresponding pixel coordinates on the image. The function signature and its return values are as follows:
\begin{lstlisting}[language=Python, caption={Doc-string for \texttt{project\_to\_pixel}}]
Args:
    pts: (*self.shape, *group_shape, 3)
    depth_is_along_ray: bool

Returns:
    pix: (*self.shape, *group_shape, pixel_dim)
    depth: (*self.shape, *group_shape)
    valid: (*self.shape, *group_shape)
\end{lstlisting}
This function returns three items:
\begin{itemize}
    \item \texttt{pix} = $p_1(x)$: The pixel coordinates corresponding to the 3D point.
    \item \texttt{depth} = $p_2(x)$: The distance value associated with the 3D point if \texttt{depth\_is\_along\_ray} is \texttt{True}; otherwise, it gives the depth, \ie, the z-component of the distance along the ray.
    \item \texttt{valid} = \texttt{True} if $x \in A$, and \texttt{False} otherwise, where $A$ is the valid projection area.
\end{itemize}

\paragraph{Function: pixel\_to\_ray}
The \texttt{pixel\_to\_ray} function performs the inverse operation, converting pixel coordinates back to a ray in 3D space. The function signature and its return values are as follows:
\begin{lstlisting}[language=Python, caption={Doc-string for \texttt{pixel\_to\_ray}}]
Args:
    pix: (*self.shape, *group_shape, pixel_dim)
    unit_vec: bool

Returns:
    origin: (*self.shape, *group_shape, 3)
    dirs: (*self.shape, *group_shape, 3)
    valid: (*self.shape, *group_shape)
\end{lstlisting}
This function also returns three items:
\begin{itemize}
    \item \texttt{origin} = $\phi_1(u)$: The origin of the ray corresponding to the pixel coordinates.
    \item \texttt{dirs} = $\phi_2(u)$: The direction vector of the ray. If \texttt{unit\_vec} is \texttt{False}, the vector is scaled such that the z-component is 1.
    \item \texttt{valid} = \texttt{True} if $u \in U$, and \texttt{False} otherwise, where $U$ is the valid pixel domain.
\end{itemize}

Example: \texttt{project\_to\_pixel}

\begin{lstlisting}[language=Python, caption={Projecting 3D Points to Pixels}]
>>> cam = cameras.OrthographicCamera.make(torch.eye(3), z_min=0.0)
>>> pts = torch.tensor([[ 1.,  2.,  5.],
                        [ 3., -2.,  8.],
                        [-2.,  3., -5.]])
>>> pixel, depth, valid = cam.project_to_pixel(pts)
>>> pixel
tensor([[ 1.,  2.],
        [ 3., -2.],
        [-2.,  3.]])
>>> depth
tensor([ 5.,  8., -5.])
>>> valid
tensor([ True,  True, False]) #-5 < z_min so last point is marked invalid
\end{lstlisting}
This example shows a scalar camera acting on a group of three points. The cameras in \textit{nvTorchCam} can operate on arbitrary groups of points and support batch-like dimensions.

Example: \texttt{pixel\_to\_ray}

\begin{lstlisting}[language=Python, caption={Converting Pixels to Rays}]
>>> cam = cameras.OrthographicCamera.make(torch.eye(3), z_min=0.0)
>>> pix = torch.tensor([[ 1., 2.],
                    [ 3., -2.],
                    [-2., 3.]])
>>> origin, dirs, valid = cam.pixel_to_ray(pix, unit_vec = False) 
>>> origin
tensor([[ 1., 2., 0.], [ 3., -2., 0.], [-2., 3., 0.]])
>>> dirs  
tensor([[0., 0., 1.], [0., 0., 1.], [0., 0., 1.]])
>>> valid
tensor([True, True, True])
\end{lstlisting}

Please refer to the source code for other useful camera functions, such as \texttt{get\_camera\_rays} and \texttt{unproject\_depth}.

\subsubsection{Tensor-like Properties}

In addition to \texttt{.shape}, camera models in \textit{nvTorchCam} support a range of tensor-like operations, such as \texttt{device}, \texttt{to}, \texttt{reshape}, \texttt{permute}, \texttt{transpose}, \texttt{squeeze}, \texttt{unsqueeze}, \texttt{expand}, \texttt{flip}, \texttt{detach}, and \texttt{clone}. Cameras also support tensor-slicing operations. Note that these operations may return either views or copies of the original camera data.

You can iterate over camera parameters using the \texttt{named\_tensors} method, which is particularly useful when setting parameters with \texttt{requires\_grad}.

\subsubsection{Heterogeneous Camera Batches}

Heterogeneous batches of cameras (batches containing different camera models) can be created using \texttt{torch.cat} and \texttt{torch.stack}:

\begin{lstlisting}[language=Python, caption={Creating Heterogeneous Camera Batches}]
>>> pin_cam = cameras.PinholeCamera.make(torch.eye(3))
>>> ortho_cam = cameras.OrthographicCamera.make(torch.eye(3))
>>> mixed_batch = torch.stack([pin_cam, ortho_cam], dim=0)
>>> type(mixed_batch)
<class 'nvtorchcam.cameras._HeterogeneousCamera'>
>>> mixed_batch.shape
torch.Size([2])
\end{lstlisting}
When slicing heterogeneous batches, the batch will devolve to a homogeneous batch if applicable:
\begin{lstlisting}[language=Python, caption={Slicing a Heterogeneous Batch}]
>>> type(mixed_batch[0])
<class 'nvtorchcam.cameras.PinholeCamera'>
>>> type(mixed_batch[1]) 
<class 'nvtorchcam.cameras.OrthographicCamera'>
\end{lstlisting}
\textbf{Note:} The \texttt{\_HeterogeneousCamera} class should not be created directly. Also, \texttt{CubeCamera} objects cannot be used in heterogeneous batches and will raise an error if concatenated with another type.

\subsubsection{Stacking in PyTorch Dataloaders}

When the library is imported, \texttt{CameraBase} objects are automatically added to PyTorch's \texttt{default\_collate\_fn\_map}. This allows cameras to be returned just like tensors in a \texttt{torch.utils.data.Dataset}'s \texttt{\_\_getitem\_\_} method, and they will be properly stacked, even if different samples use different camera models.

\subsection{Differentiable Newton Inverse}
\label{subsec:differentiable_newton_inverse}
For some camera models, such as \texttt{OpenCVCamera} and \texttt{OpenCVFisheyeCamera}, the pixel-to-ray function (often referred to as ``undistortion'') does not have a closed-form expression. Similarly to libraries like Colmap~\cite{Colmap}, we use Newton's method to compute this inverse. However, our implementation differs in two significant ways:
\begin{enumerate}
    \item We use PyTorch's autodiff framework for calculating the Jacobian in Newton's method.
    \item Our computation of the pixel-to-ray function is differentiable.
\end{enumerate}
This section focuses on the second part: making Newton's method differentiable using the inverse and implicit function theorem. We will explain the details below.

Suppose we implement a function $y = f(x; \theta)$ in PyTorch's differentiable framework. We want to implement its inverse in a differentiable manner, i.e., we aim to compute $x = g(y; \theta)$, such that:
\begin{equation}
    y = f(g(y; \theta); \theta).
    \label{fg_compose}
\end{equation}
Newton's method provides a way to calculate $g(y; \theta)$, but the question arises: how can we compute its derivatives? This can be done using the implicit and inverse function theorems. By differentiating Equation~\ref{fg_compose} with respect to $y$ and $\theta$, we get the following:
\begin{equation}
    Id = \frac{\partial f}{\partial x} \frac{\partial g}{\partial y},~~~\text{and}
\end{equation}
\begin{equation}
    0 = \frac{\partial f}{\partial x} \frac{\partial g}{\partial \theta} + \frac{\partial f}{\partial \theta}.
\end{equation}
Rearranging these equations gives:
\begin{equation}
    \frac{\partial g}{\partial y} = \left( \frac{\partial f}{\partial x} \right)^{-1},~~~\text{and}
\end{equation}
\begin{equation}
    \frac{\partial g}{\partial \theta} = - \left( \frac{\partial f}{\partial x} \right)^{-1} \frac{\partial f}{\partial \theta}.
\end{equation}
Both of these derivatives can be computed using PyTorch's \texttt{autograd} functionality, and this process is implemented in the \texttt{DifferentiableNewtonInverse} class.
The stability of this technique require further study and investigation.

\section{Image Sampling, and Cropping}


\subsection{Image Sampling}

\begin{figure*}[h!]
    \centering
    \subcaptionbox{Normalized Coordinates}[0.45\textwidth]{
        \resizebox{0.45\textwidth}{!}{
            \begin{tikzpicture}
                \clip (-1.7,-1.5) rectangle (2.5,1.5);
                
                \draw[step=0.5, thin, gray!40] (-1,-1) grid (2,1);
                \foreach \x in {-0.75,-0.25,...,1.75} {
                    \foreach \y in {-0.75,-0.25,...,1.0} {
                        \fill (\x,\y) circle (0.04);
                    }
                }

                \node at (-1.2,1.15) {\tiny $(-1,1)$};
                \node at (2,1.15) {\tiny $(1,1)$};
                \node at (-1.2,-1.15) {\tiny $(-1,-1)$};
                \node at (2,-1.15) {\tiny $(1,-1)$};

                \draw[->, thick] (-1,1.0) -- (-0.5,1.0);
                \draw[->, thick] (-1,1) -- (-1,0.5);
            \end{tikzpicture}
        }
    }
    \hspace{0.04\textwidth}  
    \subcaptionbox{Pixel Coordinates}[0.45\textwidth]{
        \resizebox{0.45\textwidth}{!}{
            \begin{tikzpicture}
                \clip (-1.7,-1.5) rectangle (2.5,1.5);
                
                \draw[step=0.5, thin, gray!40] (-1,-1) grid (2,1);
                \foreach \x in {-0.75,-0.25,...,1.75} {
                    \foreach \y in {-0.75,-0.25,...,1.0} {
                        \fill (\x,\y) circle (0.04);
                    }
                }

                \node at (-1.2,1.15) {\tiny $(0,0)$};
                \node at (2,1.15) {\tiny $(6,0)$};
                \node at (-1.2,-1.15) {\tiny $(0,4)$};
                \node at (2,-1.15) {\tiny $(6,4)$};

                \draw[->, thick] (-1,1.0) -- (-0.5,1.0);
                \draw[->, thick] (-1,1) -- (-1,0.5);
            \end{tikzpicture}
        }
    }
    \caption{Image coordinate systems}
    \label{fig:sampling_grids}
\end{figure*}
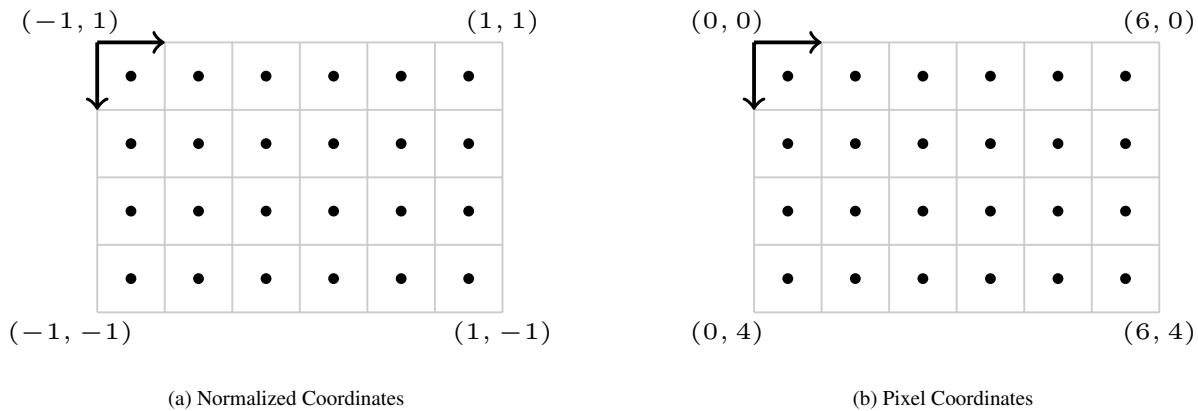

In practice, sampling locations on $U$ is necessary to represent an image on a computer. For affine cameras (see Section \ref{sec: camera_model_defs}), this is typically done using a uniform grid. Affine cameras offer flexibility in coordinate systems, with pixel coordinates and normalized coordinates being two commonly used options, as shown in Figure~\ref{fig:sampling_grids}. In \textit{nvTorchCam}, normalized coordinates are preferred because they are resolution-independent and are the required input for the \texttt{samples\_from\_images} function, a wrapper around PyTorch's \texttt{grid\_sample} that supports inferred batching. This makes them particularly useful when working with feature maps derived from images at different resolutions. For example, in CasMVSNet~\cite{CasMVSNet}, features for the same image are extracted at multiple resolutions. By using normalized coordinates, we can share camera parameters between images and their lower-resolution feature maps, ensuring consistency across different scales. We apply a similar concept for cubemaps; further details can be found in the GitHub repository.

If you have calibrated your cameras in pixel coordinates (e.g., using OpenCV~\cite{opencv}), you can normalize the intrinsic matrices with the function \lstinline|normalized_intrinsics_from_pixel_intrinsics| and revert them with \lstinline|pixel_intrinsics_from_normalized_intrinsics|.
Similarly, if you have points on an image in pixel coordinates, you can normalize them using \lstinline|pixel_coords_from_normalized_coords| and convert them back with \lstinline|normalized_coords_from_pixel_coords|.

\subsection{Cropping}

A key advantage of \textit{nvTorchCam} is the ability to keep track of how cameras change when undergoing operations like cropping or flipping.

For cropping operations, the intrinsic parameters $f_0$, $f_1$, $c_0$, $c_1$ must be adjusted accordingly. The preferred method is to use the \texttt{TensorDictionaryAffineCamera.crop} method as demonstrated below:

\begin{lstlisting}[language=Python, caption={Cropping image and camera intrinsics}]
# Define cropping dimensions as a tensor of left, right, top, bottom
>>> image_cropped = image[:, lrtb[2]:lrtb[3], lrtb[0]:lrtb[1]]
>>> cam_cropped = cam.crop(lrtb, normalized=False, image_shape=image.shape[-2:])
\end{lstlisting}

We can verify the correctness of the cropping operation by demonstrating that cropping the ``direction image'' yields the same result as cropping the camera and then generating the direction image, as shown below.

\begin{lstlisting}[language=Python, caption={Verifying camera cropping correctness}]
>>> cam = cameras.PinholeCamera.make(torch.eye(3))
>>> hw = (20,50)
>>> lrtb = torch.tensor([3,36,5,17])
>>> _, dirs, _ = cam.get_camera_rays(hw, True)
>>> dirs_cropped = dirs[lrtb[2]:lrtb[3], lrtb[0]:lrtb[1],:]
>>> cam_cropped = cam.crop(lrtb, normalized=False, image_shape=hw)
>>> _, dirs_from_cam_cropped, _ = cam_cropped.get_camera_rays(dirs_cropped.shape[:2], True)
>>> torch.testing.assert_close(dirs_cropped, dirs_from_cam_cropped)
\end{lstlisting}

Images can also be cropped directly using normalized coordinates by setting \texttt{normalized=True}. We provide utilities for random cropping, updating the intrinsics, and even sharing the crop across video frames. Flipping is supported in two modes: one that updates the intrinsics so that one focal length becomes negative, and another that flips the extrinsics, effectively flipping the entire scene geometry while keeping both focal lengths positive. For details, see \texttt{warpings.RandomResizedCropFlip}.

\subsection{Interpolation}

When evaluating points on the image, we need to interpolate the grid. This is done using \textit{samples\_from\_image} for flat images and \textit{samples\_from\_cubemap} for cubemaps. Both functions only support normalized image coordinates.

\section{Warping}
\label{sec:warping}
This section discusses backward warping and its many applications.

\subsection{General Backward Warping}

Backward warping involves transforming a source image into the perspective of a target camera using their relative pose and the target's depth map. This process allows us to synthesize the view of the source image as seen from the target camera barring occlusions and scene dynamics.

In the library, the process involves two key functions: \texttt{backward\_warp\_pts} and \texttt{backward\_warp}. Although they are presented separately, \texttt{backward\_warp\_pts} serves as a helper function for \texttt{backward\_warp} but is also useful in its own right. Now we describe each function in detail.

\paragraph{backward\_warp\_pts} handles the unprojection of points from the target camera $(U^{\text{trg}}, A^{\text{trg}}, \phi^{\text{trg}}, p^{\text{trg}})$ using the depth map $d: U^{\text{trg}} \rightarrow \mathbb{R}$, followed by reprojecting these points onto the source camera $(U^{\text{src}}, A^{\text{src}}, \phi^{\text{src}}, p^{\text{src}})$. This operation is defined by the function $E^{\text{trg} \rightarrow \text{src}}: U^{\text{trg}} \times \mathbb{R} \rightarrow U^{\text{src}} \times \mathbb{R}$ as follows:
\begin{equation}
    E^{\text{trg} \rightarrow \text{src}}(u,d) = p^{\text{src}}\left( R\left( \phi_1^{\text{trg}}(u) + d \, \phi_2^{\text{trg}}(u) \right) + T \right).
    \label{eq:backward_warp_pts}
\end{equation}
Here, $R$ and $T$ represent the relative rotation and translation between the target and source cameras. We denote this function as $E$ for ``epipolar'' because, when $u$ is fixed and $d$ varies, $E(u,d)$ traces the epipolar curve in the source image corresponding to the point $u$ in the target image.

\paragraph{backward\_warp}
The function \texttt{backward\_warp} performs the actual sampling of the source image at the points computed by \texttt{backward\_warp\_pts}. Formally, this can be expressed as:
\begin{equation}
    I^{\text{src} \rightarrow \text{trg}}(u,d) = I^{\text{src}}( E^{\text{trg} \rightarrow \text{src}}_1(u,d) ).
\end{equation}
In this equation, $I^{\text{src}}$ denotes the source image, and $I^{\text{src} \rightarrow \text{trg}}$ is the warped source image viewed from the target camera’s perspective.

\subsection{Cost-Volume}

A key application of backward warping is the construction of cost volumes, such as in plane or sphere sweeping networks. This process involves sampling multiple depth (plane sweeping) or distance (sphere sweeping) hypotheses and backward warping one or more source images to align with a target image based on these hypotheses. At the correct hypothesis, the warped source images will accurately align with the target. Some works, such as~\cite{CasMVSNet,360MVSNet}, have extended this by using pixel-wise hypotheses, which are also supported by the \texttt{backward\_warp} function.

\begin{figure*}[h]
    \centering
    \subfloat[Reference image]{\includegraphics[width=0.32\textwidth]{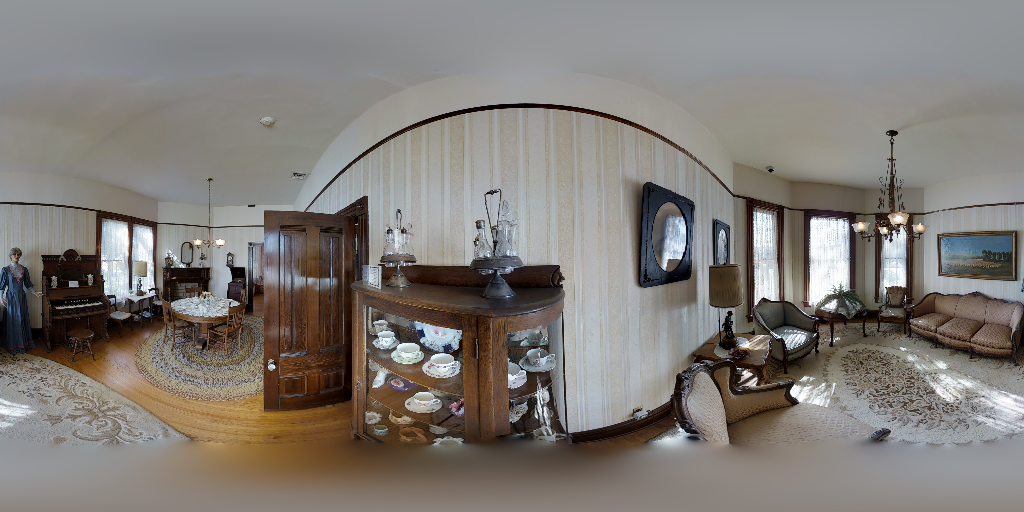}}~
    \subfloat[Source warped at distance 1.35]{\includegraphics[width=0.32\textwidth]{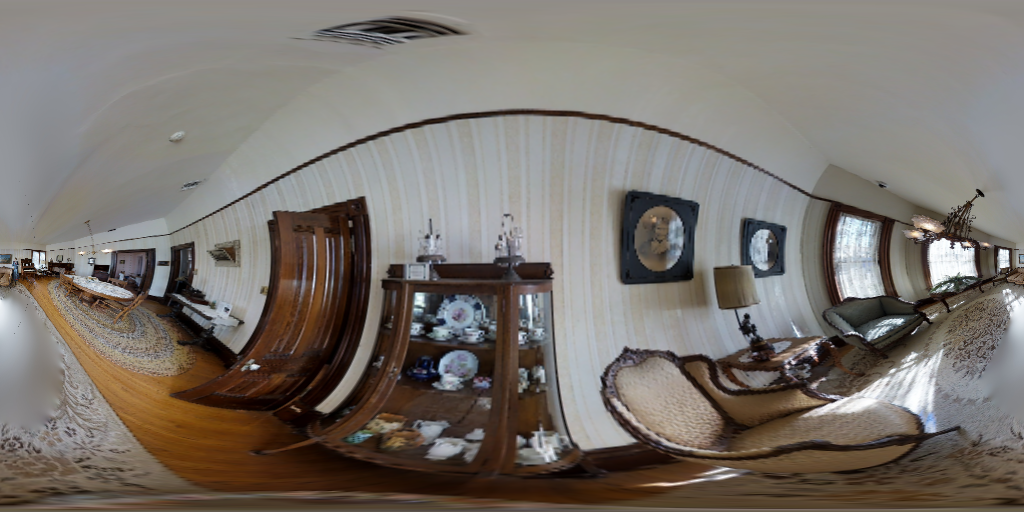}}~
    \subfloat[Source warped at distance 1.6]{\includegraphics[width=0.32\textwidth]{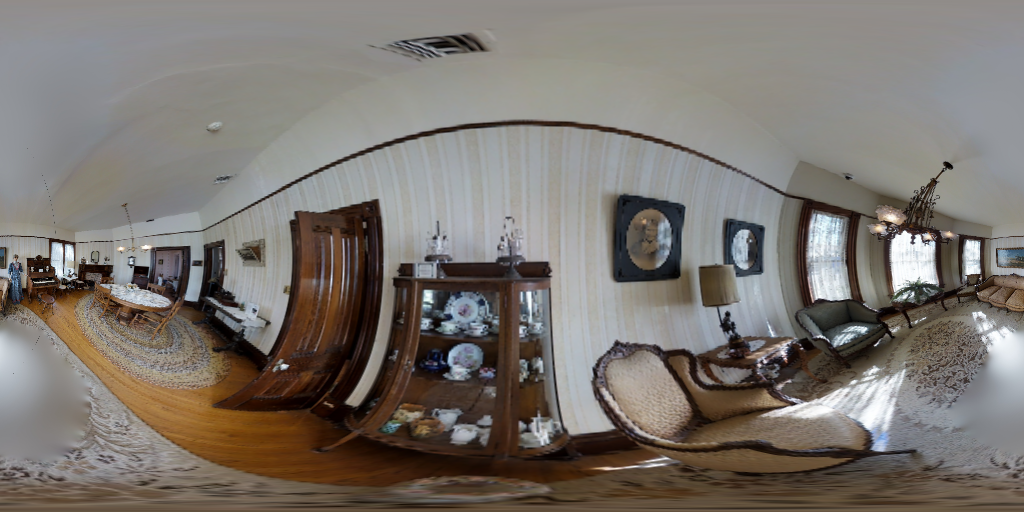}}\\
    \subfloat[Source image]{\includegraphics[width=0.32\textwidth]{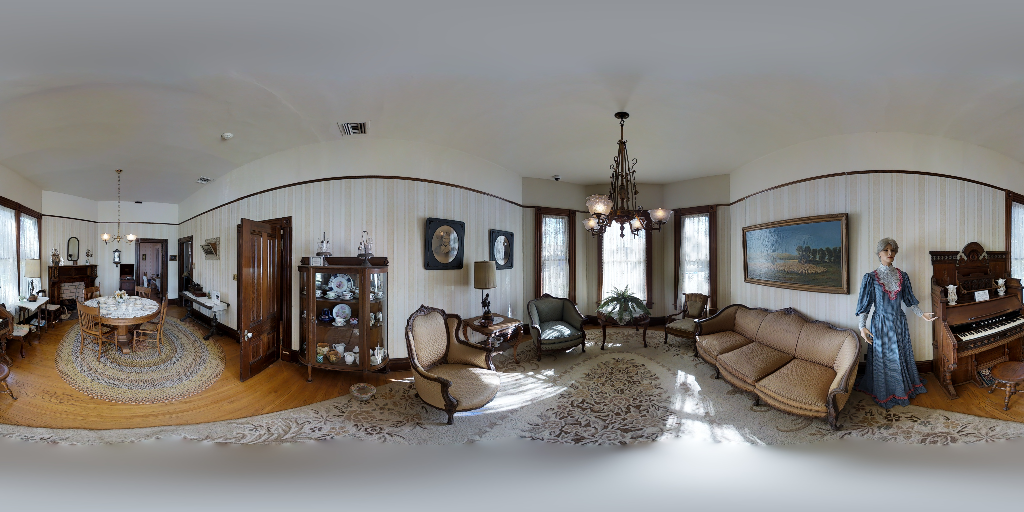}}~
    \subfloat[Source warped at distance 2.13]{\includegraphics[width=0.32\textwidth]{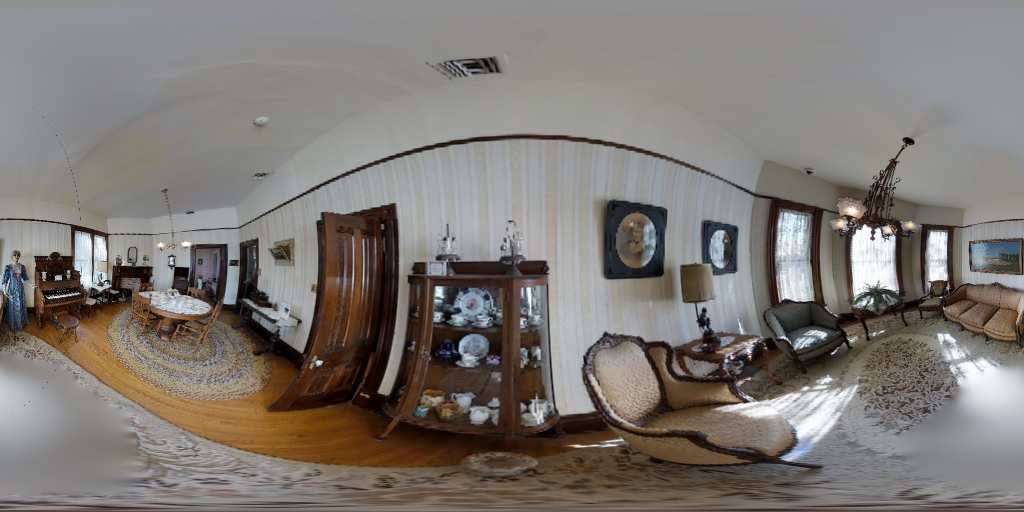}}~
    \subfloat[Source warped at distance 3.61]{\includegraphics[width=0.32\textwidth]{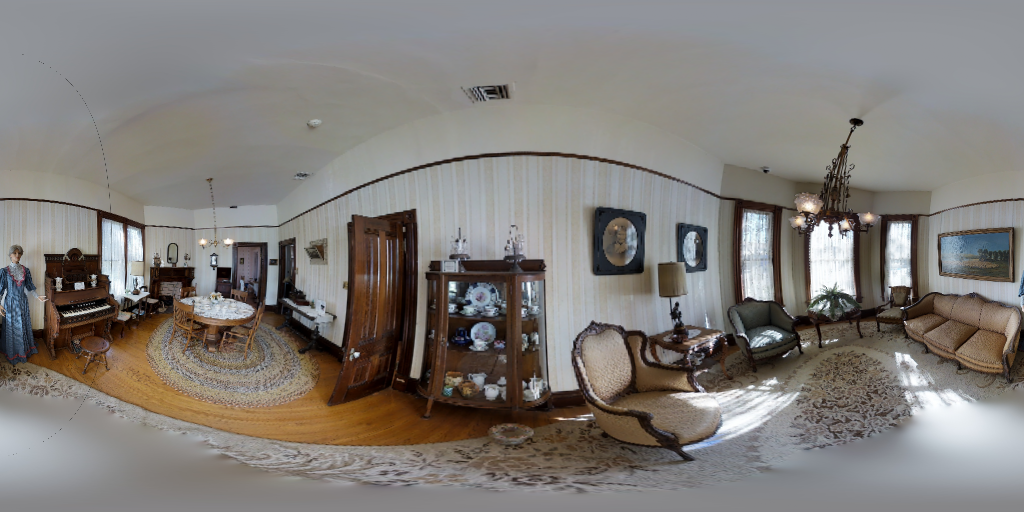}}
    \caption{Example of the result of warping the source image at different distances from the camera that captured the reference image.} 
    \label{fig:cost_volume}
\end{figure*}

All of this functionality is implemented in a camera-independent manner within \textit{nvTorchCam}, and the high level of vectorization through inferred batching allows for efficient warping of multiple source images based on multiple hypotheses, without requiring explicit loops. Figure~\ref{fig:cost_volume} illustrates the sphere sweeping process for ERP images.

\subsection{Resample by Intrinsics}

For central cameras, it can be shown that backward warping with any depth or distance map can be used to warp one camera model to another~\cite{Fova-depth}. Varying $R$ is equivalent to applying an extrinsic rotation to the camera. This process is implemented in the \texttt{resample\_by\_intrinsics} function, which uses a constant distance map set to one. Other applications include splitting a fisheye image into multiple pinhole images or dividing an ERP into pinholes on an icosahedron. Figure~\ref{fig:resample_by_intrinsics} illustrates the process of warping an ERP to other models, potentially with an additional extrinsic rotation.

\subsection{Consistency Checking and MVS Fusion}
\label{subsec:consistency_checking}

\begin{figure}[h!]
    \resizebox{0.45\textwidth}{!}{
        \begin{tikzpicture}
        \input{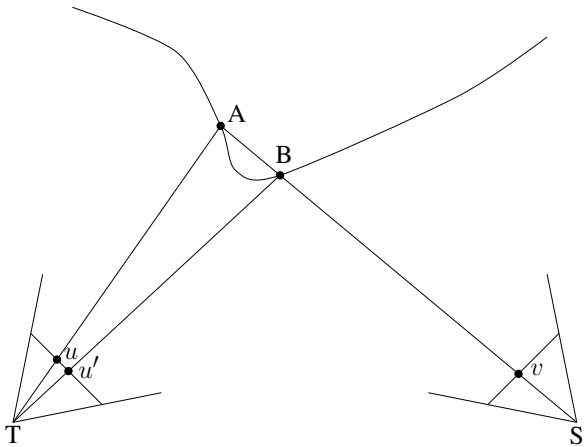}
        \end{tikzpicture}
    }
    \caption{Illustration for Subsection~\ref{subsec:consistency_checking}}
    \label{fig:consistency_checking}
\end{figure}

Another important application of backward warping is checking consistency between depth maps from different views. This technique is useful for identifying occlusions and for MVS-Fusion, which involves fusing predicted depths from multi-view stereo (MVS) estimates.

Our MVS-Fusion technique, implemented in the function \texttt{fuse\_depths\_mvsnet}, is an adaptation of the method implemented in~\cite{MVSFiltering}. It has been modified to support arbitrary camera models, run efficiently on the GPU, and is fully vectorized to eliminate the need for loops.

Next, we explain how this works. Please refer to Figure~\ref{fig:consistency_checking} to better understand the geometry reflected by the following equations. First, assume we have source and target cameras and their corresponding depth map estimates $D^{\text{src}}$ and $D^{\text{trg}}$. We first unproject the pixel $u$ in the target camera using its depth $D^{\text{trg}}(u)$ and reproject it onto the source camera
\begin{equation}
    v = E_1^{\text{trg} \rightarrow \text{src}}(u, D^{\text{trg}}(u)).
\end{equation}
We then unproject $v$ using $D^{\text{src}}(v)$ and reproject it onto the target
\begin{equation}
    u', \hat{D}^{\text{trg}}(u) = E^{\text{src} \rightarrow \text{trg}}(v, D^{\text{src}}(v)), 
\end{equation}
where $u'$ is the new location and $\hat{D}^{\text{trg}}(u)$ the corresponding depth.
In Figure~\ref{fig:consistency_checking}, $D^{\text{src}}(v)$ corresponds to the length from $S$ to $B$, and  $\hat{D}^{\text{trg}}(u)$ corresponds to the length from $T$ to $B$.
Note that $\hat{D}^{\text{trg}}(u)$ is an estimate of the depth in the target based on the source estimate.

\begin{figure*}[h]
    \centering
    \subfloat[]{\includegraphics[width=0.49\textwidth,keepaspectratio]{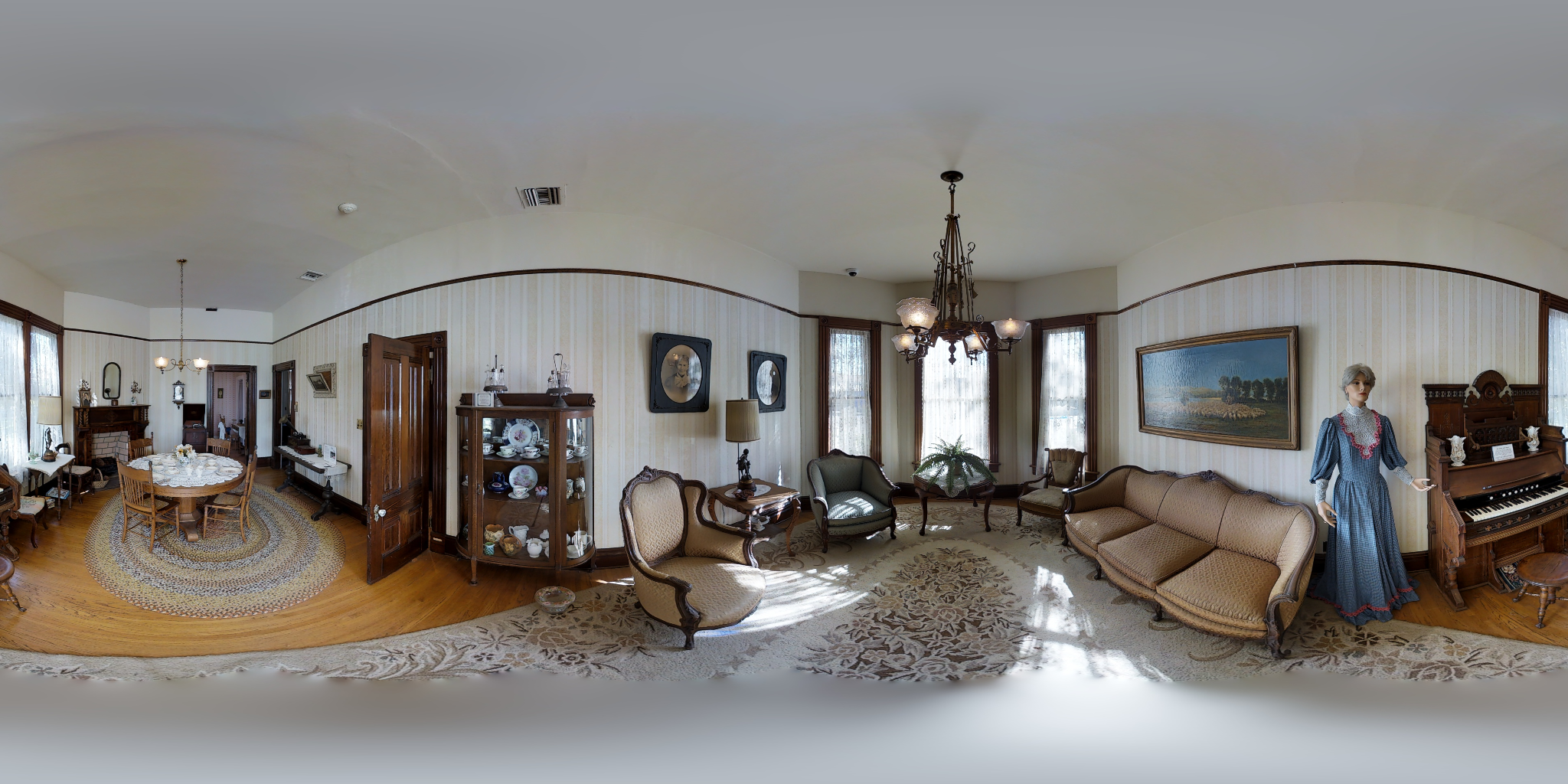}}~
    \subfloat[]{\includegraphics[width=0.49\textwidth,keepaspectratio]{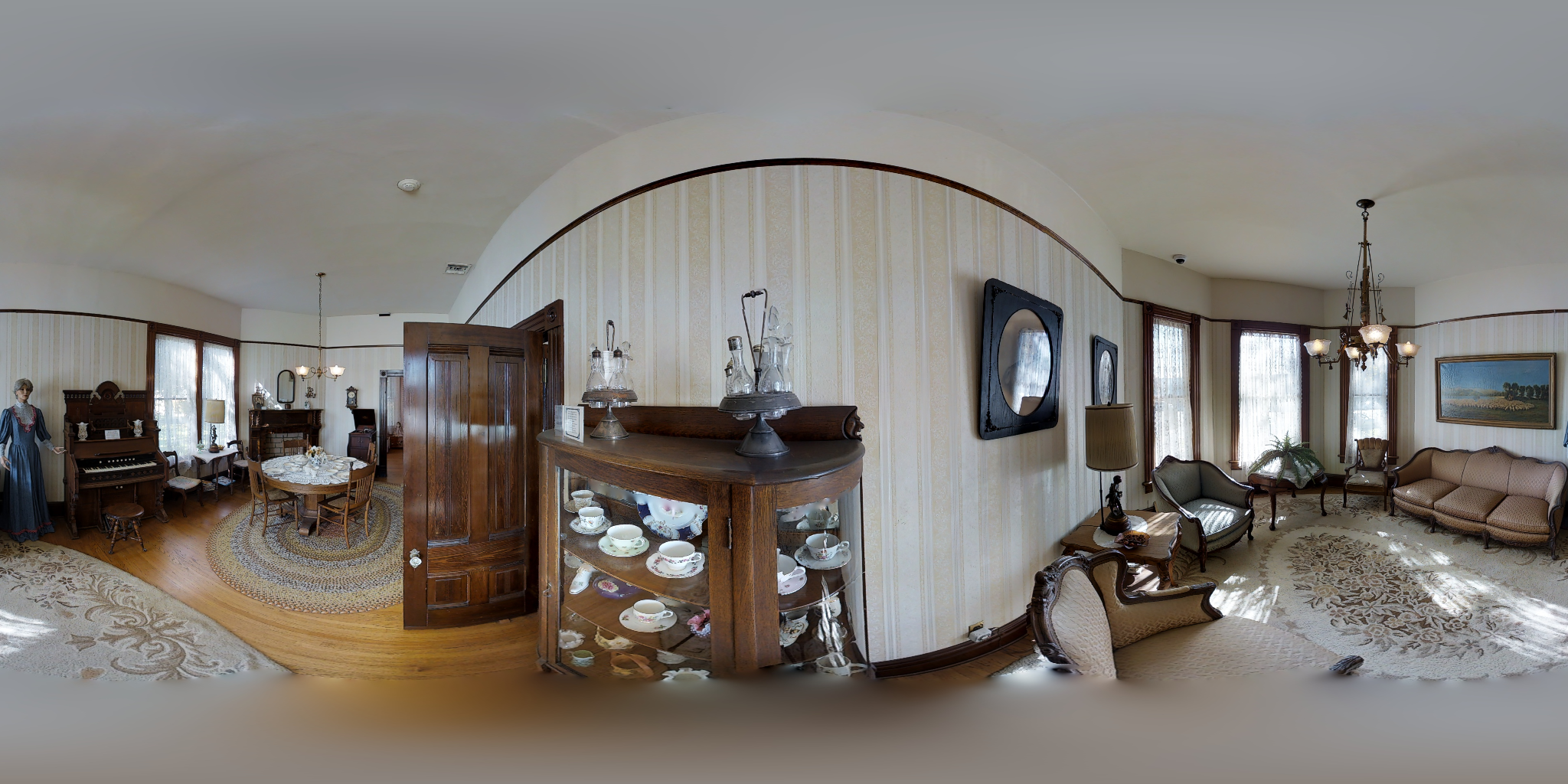}}\\
    \subfloat[]{\includegraphics[width=0.49\textwidth,keepaspectratio]{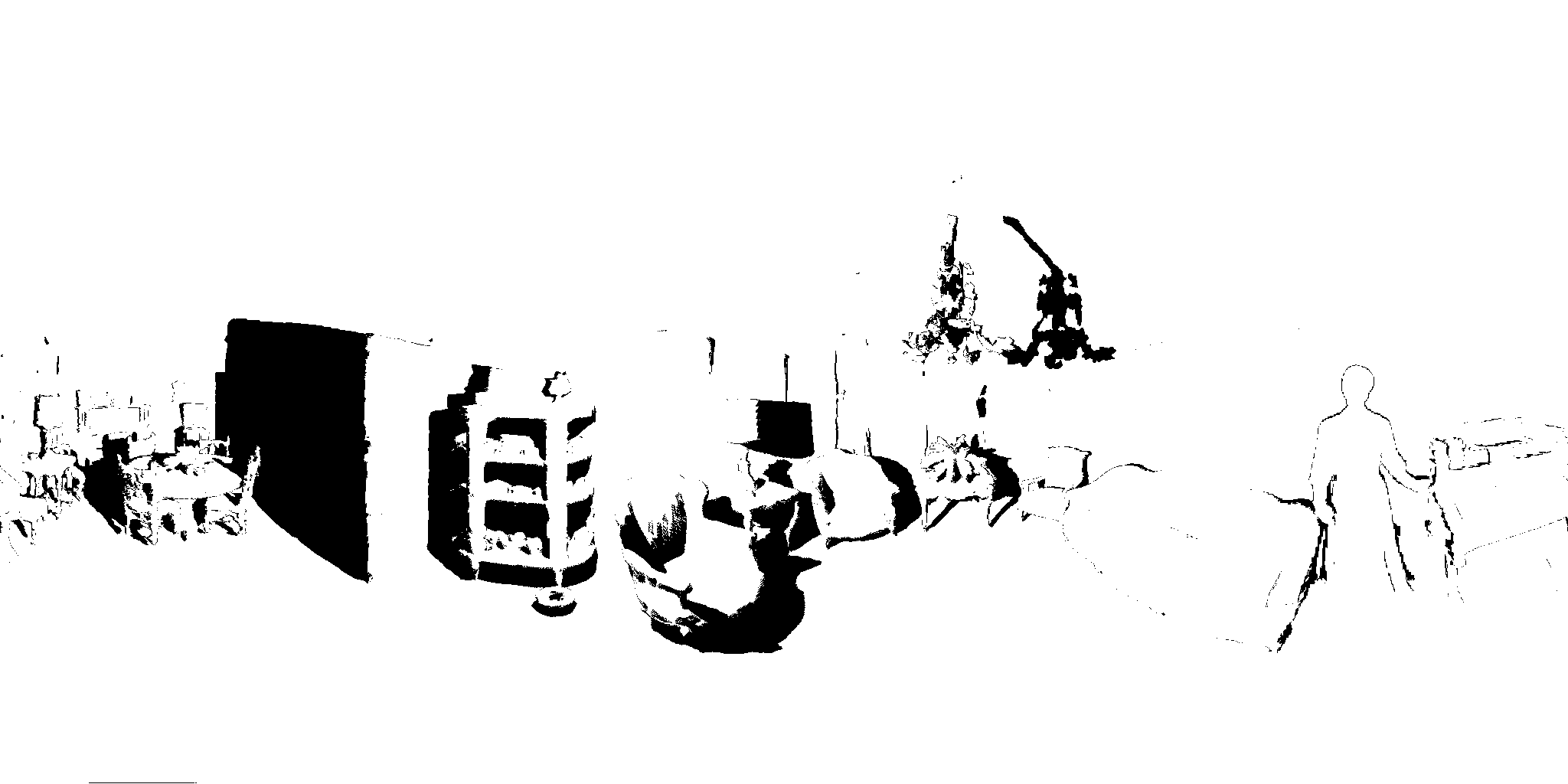}}~
    \subfloat[]{\includegraphics[width=0.49\textwidth,keepaspectratio]{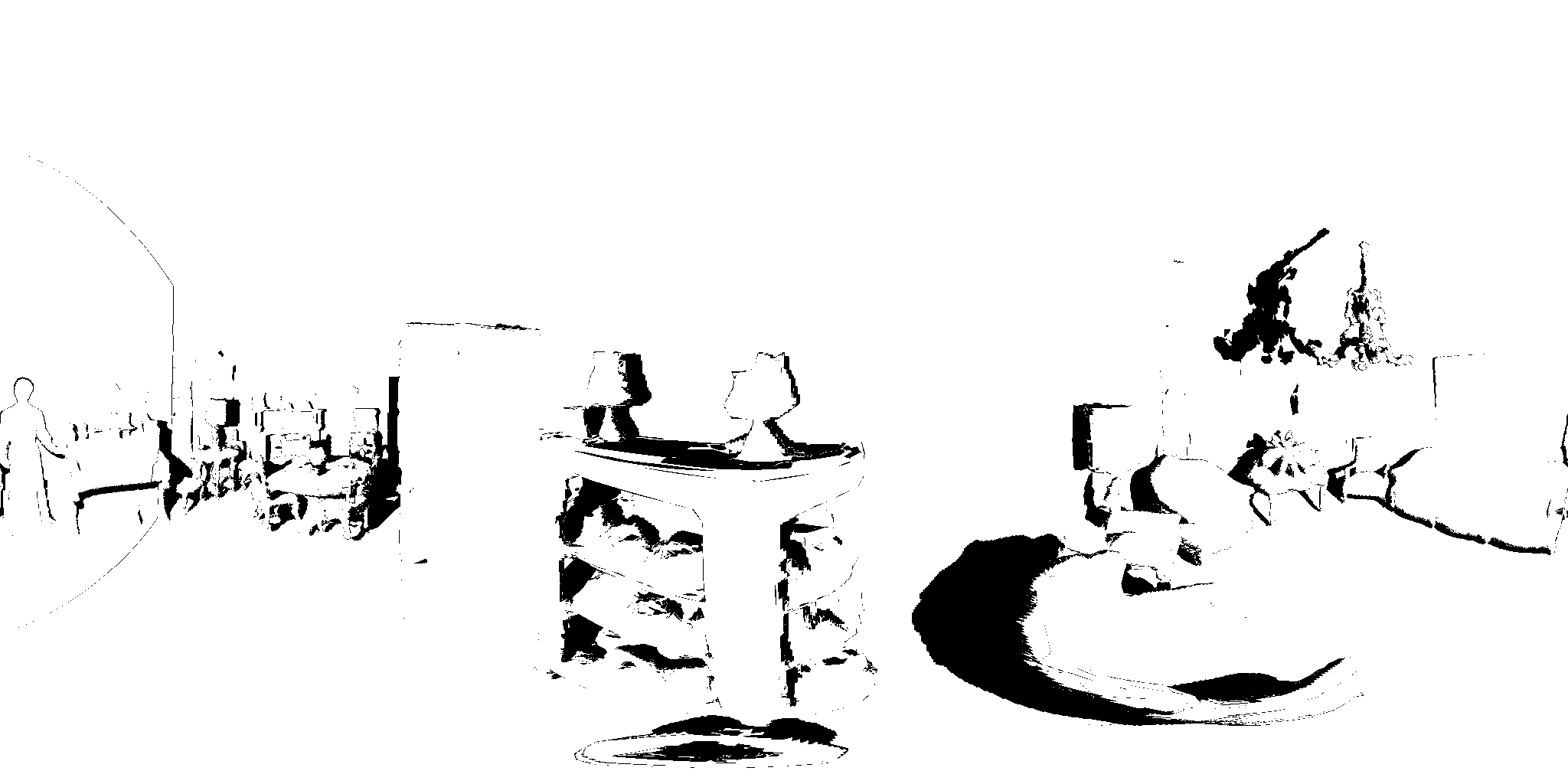}}
    \caption{Mutual visibility masks calculated by backward-forward warping. (c) is white at locations in (a) that are visible in (b). (d) is white at locations in (b) that are visible in (a)}
    \label{fig:mutual_visibility}
\end{figure*}

We can then check the consistency between the pixels $u$ and $u'$ and their corresponding depths $\hat{D}^{\text{trg}}(u)$ and $D^{\text{trg}}(u)$.
We create a mask that is 1 if $u$ and $u'$ are within a spatial threshold $\tau_1$ and if their relative depths are within a threshold $\tau_2$:
\begin{equation}
    \text{mask}^{\text{src}}(u)=
    \begin{cases}
    1 & \left\| u - u' \right\| < \tau_1 \land \frac{\left|\hat{D}^{\text{trg}}(u) - D^{\text{trg}}(u)\right|}{D^{\text{trg}}(u)} < \tau_2\\
    0 &\text{Otherwise}
    \end{cases}.
\end{equation}
An example of such masks is shown in Figure \ref{fig:mutual_visibility}.
If we have multiple source views, we can fuse the depth estimates from the sources $\hat{D}^{\text{src}}(u)$ where they are consistent with the source depth $D^{\text{src}}$, based on the consistency masks.

If there are multiple sources $D^{\text{src}_i}$ for $i = \{1:N\}$, they can be fused with the target based on the masks:
\begin{equation}
    \text{fused\_depth}^{\text{trg}}(u) = \frac{\sum_{i=0}^{N} \text{mask}^{\text{src}_i}(u) \cdot \hat{D}^{\text{src}_i}(u)}{\sum_{i=0}^{N} \text{mask}^{\text{src}_i}(u)},
\end{equation}
where we let $D^{\text{src}_0} = D^{\text{trg}}$ to simplify the notation.

\subsection{Stereo Rectification (including ERPs)}

A key application of the \texttt{resample\_by\_intrinsics} function is stereo rectification, where a pair of images from central cameras is warped, or extrinsically rotated, to appear as if taken side-by-side, ensuring that the epipolar lines are horizontal. Similarly, \cite{Li_Jin2022MODE} proposes rectifying a pair of images onto equirectangular projections (ERPs) so they appear as if taken on top of each other, making the epipolar curves vertical. We observe that these two processes are fundamentally similar, and both are implemented in the \texttt{stereo\_rectify} function.

In this discussion, we explain the ``on-top'' ERP rectification case, i.e., the case where the epipolar curves are vertical. The ``side-by-side'' case is analogous, with the primary difference being that the y-axis is replaced by the x-axis. For more details, the implementation in \texttt{stereo\_rectify} provides further clarification.

The core task in rectification is finding the appropriate extrinsic rotation matrix to align the cameras on top of each other. Given the initial extrinsics $(R^i, T^i)$ for cameras $i \in \{0, 1\}$, where $R^i$ is the rotation matrix composed of column vectors $R^i = [r^i_0, r^i_1, r^i_2]$, aligning the cameras ``on top of each other'' requires that $r^0_1 = r^1_1 = \frac{T^1 - T^0}{||T^1 - T^0||}$, and $R^0 = R^1$.

\begin{figure*}[h]
    \centering
    \begin{tabular}{cc}
        \includegraphics[width=0.45\textwidth,keepaspectratio]{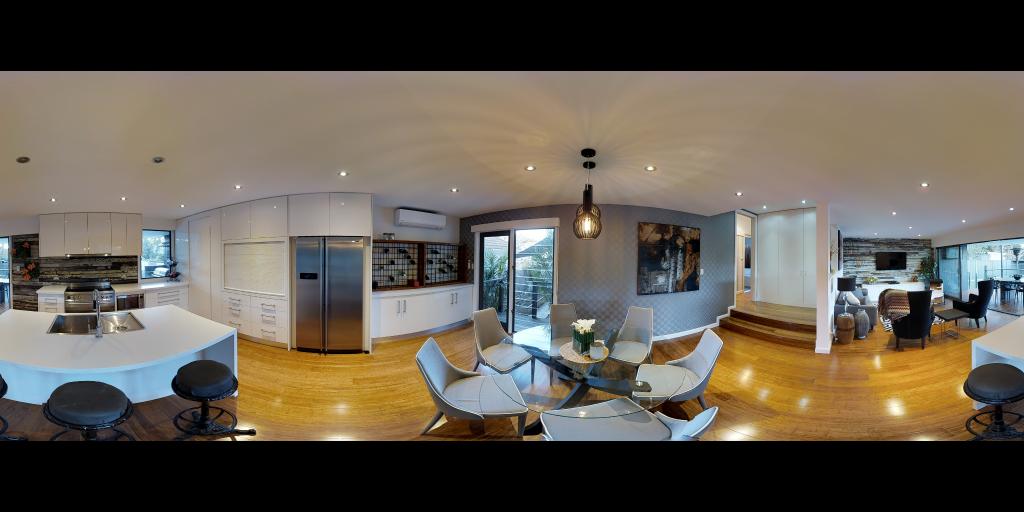} & \includegraphics[width=0.45\textwidth,keepaspectratio]{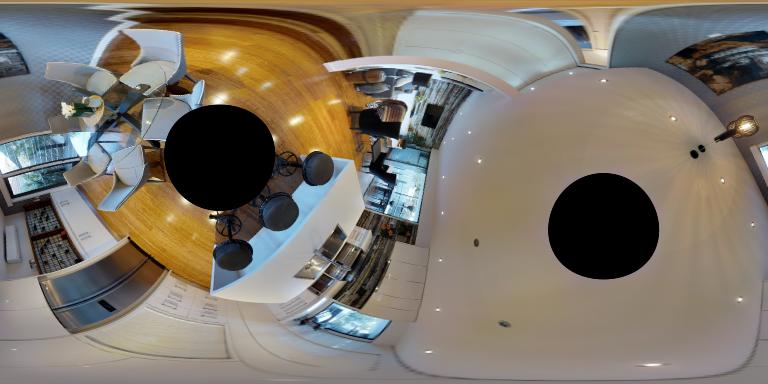} \\
        \includegraphics[width=0.45\textwidth,keepaspectratio]{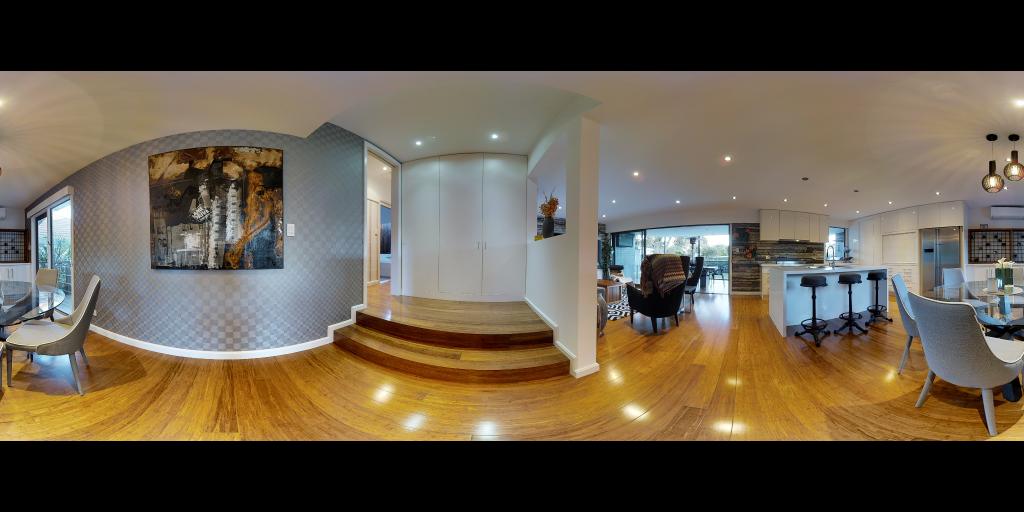} & \includegraphics[width=0.45\textwidth,keepaspectratio]{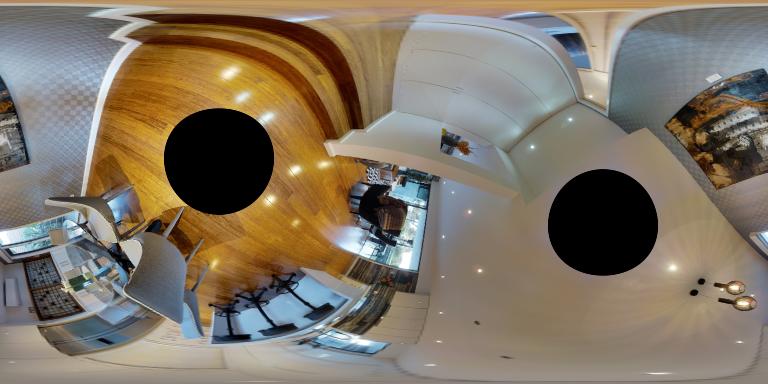} \\
    \end{tabular}
    \caption{Original ERP images (left) and rectified ERPs such that feature matches are directly on top of each other (right).} 
    \label{fig:mode_rectification}
    
\end{figure*}

We compute the extrinsic rotation as follows: First, define $\hat{r}_1 = \frac{T^1 - T^0}{||T^1 - T^0||}$. For the on-top ERP case, we can select any vector orthogonal to $\hat{r}_1$ as $\hat{r}_2$, since the choice corresponds to a circular translation of the ERPs. Although this is mathematically valid for the pinhole side-by-side case, an arbitrary choice may cause distortions. To minimize these distortions, we choose $\hat{r}_2$ as the component of $r^0_2 + r^1_2$ that is orthogonal to $\hat{r}_1$, as this keeps the camera orientations close to their original direction. Then, define $\hat{r}_0 = \hat{r}_1 \times \hat{r}_2$ (where $\times$ denotes the cross product). The resulting matrix $\hat{R} = [\hat{r}_0, \hat{r}_1, \hat{r}_2]$ represents the desired orientation for the rectified cameras. Finally, we apply the \textit{resample\_by\_intrinsics} function to both images using the extrinsic rotations $(R^i)^{-1}\hat{R}$.
Figure \ref{fig:mode_rectification} illustrates the rectification process for two ERPs.

\section{Conclusion}
In this paper, we introduced \textit{nvTorchCam}, an open-source library designed to abstract camera models and enable differential camera operations in computer vision. Built on PyTorch, \textit{nvTorchCam} can be easily integrated into existing deep learning vision projects, simplifying the handling of various camera models and providing robust backward warping capabilities across different camera types.

Looking ahead, there are several possibilities for future development based on community interest. While the library currently supports non-central cameras, such as orthographic models, this feature has not been extensively exploited. There is potential to expand support for more non-central models like rolling shutter~\cite{Oth2013RollingShutter} and CAHVORE~\cite{mrcal}. Additionally, a unified calibration framework using PyTorch optimization could be explored to enhance camera parameter adjustments. Future developments will depend on feedback and contributions from the open-source community. 
{
    \small
    \bibliographystyle{ieeenat_fullname}
    \bibliography{main}
}

\appendix 

\section{Camera Projection Definitions}
\label{sec:projection_details}

In this section, we provide the mathematical details of each camera projection function. Note that all cameras, except the CubeCamera, are affine cameras. This means they are subclasses of \texttt{TensorDictionaryAffineCamera} and have affine parameters $f_0, f_1, c_0, c_1$.

We denote the 3D point to be projected as $(x, y, z)$ and the point projected onto the pixel as $(u, v) = p(x, y, z)$ for affine cameras and $(u, v, w) = p(x, y, z)$ for the CubeCamera.

We also discuss how the ``inverse'' \texttt{pixel\_to\_ray} function is computed; for full details, please refer to the code.

\subsection{PinholeCamera}
Parameters: $f_0, f_1, c_0, c_1$. The projection function is given by:
\[
u', v' = \frac{x}{z}, \frac{y}{z}
\]
\[
u, v = f_0 u' + c_0, f_1 v' + c_1
\]
The inverse is analytic.

\subsection{OrthographicCamera}
Parameters $f_0, f_1, c_0, c_1$. The projection function is given by:
\[
u', v' = x, y
\]
\[
u, v = f_0 u' + c_0, f_1 v' + c_1
\]
The inverse is analytic.

\subsection{OpenCVCamera}
Parameters $f_0, f_1, c_0, c_1$, and distortion coefficients $k_0, k_1, k_2, k_3, k_4, k_5, p_0, p_1$. The projection function is given by:
\[
x', y' = \frac{x}{z}, \frac{y}{z}
\]
\[
r^2 = {x'}^2 + {y'}^2
\]
\[
D_{radial} = \frac{1 + k_0 r^2 + k_1 r^4 + k_2 r^6}{1 + k_3 r^2 + k_4 r^4 + k_5 r^6}
\]
\[
u' = x' \cdot D_{radial} + 2 p_0 x' y' + p_1 (r^2 + 2 {x'}^2)
\]
\[
v' = y' \cdot D_{radial} + p_0 (r^2 + 2 {y'}^2) + 2 p_1 x' y'
\]
\[
u, v = f_0 u' + c_0, f_1 v' + c_1
\]
The inverse requires Newton's method to convert from $u', v'$ back to $x', y'$.

\subsection{EquirectangularCamera}

Parameters $f_0, f_1, c_0, c_1$. The projection function is given by:
\[
r = \sqrt{x^2 + y^2 + z^2}
\]
\[
u' = \arccos\left(-\frac{y}{r}\right)
\]
\[
v' = \text{atan2}(x, z)
\]
\[
u, v = f_0 u' + c_0, f_1 v' + c_1
\]

The inverse is analytic. 


\subsection{OpenCVFisheyeCamera}
Parameters $f_0, f_1, c_0, c_1$, and distortion coefficients $k_0, k_1, k_2, k_3$. The projection function is given by:
\[
x', y', z' = \frac{x, y, z}{\sqrt{x^2 + y^2 + z^2}}
\]
\[
\theta = \arccos(z')
\]
\[
\theta_d = \theta \left(1 + k_0 \theta^2 + k_1 \theta^4 + k_2 \theta^6 + k_3 \theta^8\right)
\]
\[
x'', y'' = \frac{x', y'}{\sqrt{x'^2 + y'^2}}
\]
\[
u', v' = \theta_d \cdot (x'', y'')
\]
\[
u, v = f_0 u' + c_0, f_1 v' + c_1
\]
The inverse requires Newton's method to convert $\theta_d$ to $\theta$.

\subsection{BackwardForwardPolynomialFisheyeCamera}
Parameters $f_0, f_1, c_0, c_1$ and polynomial coefficients $p_0, p_1, \dots, p_N$ for forward projection, and $q_0, q_1, \dots, q_M$ for backward projection. The projection function is:
\[
x', y', z' = \frac{x, y, z}{\sqrt{x^2 + y^2 + z^2}}
\]
\[
\theta = \arccos(z')
\]
\[
\theta_d = p_0 + p_1 \theta + p_2 \theta^2 \dots + p_N \theta^N
\]
\[
x'', y'' = \frac{x', y'}{\sqrt{x'^2 + y'^2}}
\]
\[
u', v' = \theta_d \cdot (x'', y'')
\]
\[
u, v = f_0 u' + c_0, f_1 v' + c_1
\]
The inverse is calculated using a ``backward polynomial'':
\[
\theta = q_0 + q_1 \theta_d + q_2 \theta_d^2 \dots + q_M \theta_d^M
\]
The backward polynomial must be specified by the user and should approximate the inverse of the forward polynomial.

\subsection{Kitti360FisheyeCamera}
Parameters $f_0, f_1, c_0, c_1$, and distortion coefficients $k_0, k_1$, and $x_i$. The projection function is:
\[
x', y', z' = \frac{x, y, z}{\sqrt{x^2 + y^2 + z^2}}
\]
\[
x'', y'' = \frac{x'}{z' + x_i}, \frac{y'}{z' + x_i}
\]
\[
r^2 = {x''}^2 + {y''}^2
\]
\[
r_d^2 = 1 + k_0 r^2 + k_1 r^4
\]
\[
u', v' = r_d^2 \cdot x'', r_d^2 \cdot y''
\]
\[
u, v = f_0 u' + c_0, f_1 v' + c_1
\]
The inverse requires Newton's method for $r^2$ back to $r_d^2$, and the rest is analytic.

\subsection{CubeCamera}
The cube camera model does not have traditional parameters. The projection function is:
\[
u, v, w = \frac{x, y, z}{\|x, y, z\|_{\infty}}
\]
if \texttt{depth\_is\_along\_ray} is set to False, or
\[
u, v, w = \frac{x, y, z}{\|x, y, z\|_2}
\]
if \texttt{depth\_is\_along\_ray} is set to True. The inverse is analytic.

\end{document}